# A New 3D Segmentation Technique for QCT Scans of the Lumbar Spine to Determine BMD and Vertebral Geometry

Andre Mastmeyer, Klaus Engelke, Christina Fuchs, Willi A. Kalender, Institute of Medical Physics, Erlangen, Germany

## Abstract

Quantitative computed tomography (QCT) is a standard method to determine bone mineral density (BMD) in the spine. Traditionally single 8 – 10 mm thick slices have been analyzed only. Current spiral CT scanners provide true 3D acquisition schemes allowing for a more differential BMD analysis and an assessment of geometric parameters, which may improve fracture prediction. We developed a novel 3D segmentation approach that combines deformable balloons, multi seeded volume growing, and dedicated morphological operations to extract the vertebral bodies. An anatomy-oriented coordinate system attached automatically to each vertebra is used to define volumes of interest. We analyzed intra-operator precision of the segmentation procedure using abdominal scans from 10 patients (60 mAs, 120 kV, slice thickness 1mm, B40s, Siemens Sensation 16). Our new segmentation method shows excellent precision errors in the order of < 1 % for BMD and < 2 % for volume.

## 1 Introduction

**Aim.** Low BMD is one of the most important risk factors for osteoporotic fractures. Quantitative computed tomography (QCT) is a standard technique to analyze trabecular BMD of the vertebrae L1-L3. Tradionally single slices of the mid-vertebral sections were scanned and analyzed. Trabecular and cortical VOIs were defined relative to an anatomy oriented coordinate system [1]. In this contribution we present a new three-dimensional volumetric approach using spiral CT scans to extract the vertebral geometry.

**State of the Art.** For volumetric BMD measurements of the spine Lang et al. [2] reported precision errors of 1.3 % although they did not use a full 3D segmentation technique. Another semi-automatic approach has been recently presented by Kaminsky et. al [3] who manually separated the individual vertebral levels before segmentation. However, this is a lengthy procedure that took 1-2 hours per dataset. Previously we developed a semi-automatic 3D segmentation for the proximal femur [4]. Due to the higher degree of complexity in the spine a simple adaptation of this approach for the spine was not very successful. Therefore in this contribution we developed fast explicit deformable models as a major component to segment the vertebral body.

## 2 Material and Methods

### 2.1 Segmentation

**Constraints.** Initially the user has to mark the centers of the vertebral bodies. In a second step the central curve of the spinal canal is detected by a rolling ball procedure [3]. Then planes are automatically fitted to the spaces of the vertebral disks. These three pre-processing steps generate a number of landmarks that are used to define disjoint enclosing cylinders for each vertebra. The planes cutting through the vertebral disks limit the extension of the cylinders along the spine. The cylinders effectively exclude the aorta.

**Balloon Segmentation.** The balloon model chosen here consists of an adaptive triangle mesh. The algorithm approximately solves the simplified Euler-Lagrangian equation of motion

$$m \ddot{\vec{p}} = \vec{f}_{smg} + \vec{f}_{img}$$

which is coupled to a minimization problem of internal and external energy. The internal energy $\vec{f}_{smg}$ depends on the connectivity forces, the external energy $\vec{f}_{img}$ is determined along profiles that radially emerge from the balloon surface [5]. The periosteal edges of the cortical shell attract the vertices of the balloon while the internal forces try to pertain a smooth shape. The procedure is able to overcome difficulties due to noise and can bridge gaps caused by very thin contours or low contrast. During the balloon growth new vertices are inserted in stretched areas. Due to the smoothness term that is a major regularizing part of a deformable model the surface is not able to catch all points in highly curved regions.

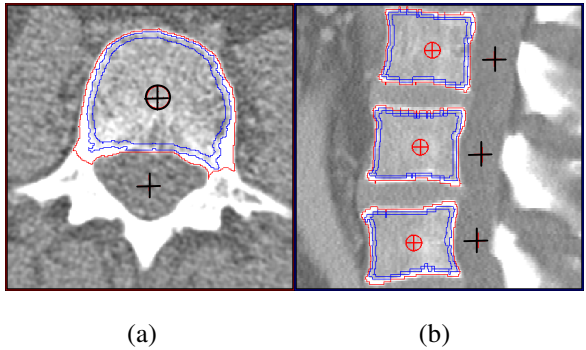

**Fig. 1** Segmentation results: a) axial, b) lateral view

**Optimization of Segmentation.** Voxels with high gray values on the balloon surface are used as seeds for volume growing operations which are followed by a closing and hole-filling procedure described in [4]. Afterwards the processes have to be separated reproducibly from the vertebral body. We use the smallest dissecting surface through the pedicles between the vertebral body and the transverse processes. It is found by subsequent erosions that result in separate residuals of the anatomical components (body and processes). The desired surface is found by non-intersecting parallel dilations of these residuals. The result is shown as outer contour in Fig. 1. Finally local erosions of voxels with high grey values determine the trabecular compartment. A second homogeneous erosion peels off the sub cortical bone (inner contours, see Fig. 1).

**Vertebral Coordinate System (VCS).** Four landmarks are used for the definition of a right handed vertebral coordinate system (VCS). Its origin (O) is defined by the center of volume (COV) of the vertebral body. Next, the central line of the vertebral column is approximated by an interpolating spline curve of the COVs of contiguous segmented vertebral bodies. The tangent to this curve in O defines the z-axis of the VCS. The y-axis is defined by the line perpendicular to the z-axis that connects O with the central line of the spinal canal. The x-axis is normal to the other two. The VCS is used to reproducibly define analysis volumes of interest (VOIs).

## 2.2 Evaluation

**Intra-operator precision** was analyzed using clinical routine abdominal scans from 10 patients. CT acquisition was performed on a Siemens Sensation 16 (60 mAs, 120 kV, slice thickness 1 mm). For each patient tomographic images with three different fields of view (FOV) (150, 250, 350 mm) were analyzed three times by the same operator. Precision errors were determined for BMD and volume of the vertebral bodies L1-L3 as follows. For each patient results for L1-L3 from each VOI were averaged. Then averages were calculated for the three analyzes resulting in a %CV per patient. Finally root mean square averages for all 10 patients were calculated resulting in a percent coefficient of variation (%CV) and a corresponding standard deviation SD per VOI.

## 3 Results

**Precision.** The intra-operator results are shown in Fig. 2. Precision errors are given for trabecular BMD and total volume. The precision errors for BMD and volume for the total segmentation are below 1 % and 2 % respectively, which is an excellent result.

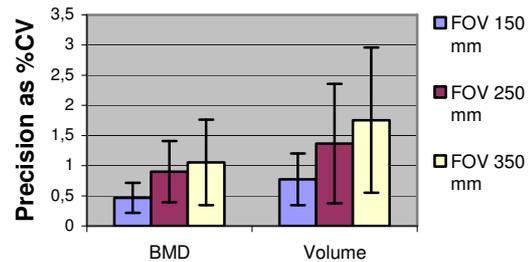

**Fig. 2** Precision errors of the total segmentation

## 4 Conclusion

The analysis of the new 3D segmentation method demonstrated comparable precision errors to existing volumetric methods [2] without 3D segmentation. We have shown that the combination of several steps corrects deficiencies of the individual methods. Improved performance should result in superior diagnostic performance, which is very important for fracture prediction in osteoporosis.